%% file: main.tex
\documentclass[runningheads]{llncs}

\usepackage{eccv}

\usepackage{eccvabbrv}
\usepackage{graphicx}
\usepackage{booktabs}
\usepackage{multirow}
\usepackage{colortbl}
\usepackage{xcolor}
\usepackage{makecell}
\usepackage{mathtools}
\usepackage[accsupp]{axessibility}  
\usepackage{float}  
\usepackage{wrapfig}

\usepackage{hyperref}

\usepackage{orcidlink}

\begin{document}

\title{Holistic Inverse Rendering of Complex Facade via Aerial 3D Scanning} 


\author {
    Zixuan Xie*\inst{1,3},
    Rengan Xie*\inst{2},
    Rong Li\inst{3},
    Kai Huang\inst{1,3},
    Pengju Qiao\inst{1,3},
    Jingsen Zhu\inst{2},
    Xu Yin\inst{4},
    Qi Ye\inst{2},
    Wei Hua\inst{3},
    Yuchi Huo\inst{2,3},
    Hujun Bao\inst{2,3}
}
\authorrunning{Xie et al.}

\institute{Institute of Computing Technology, Chinese Academy of Sciences \and
Zhejiang University \and
Zhejianglab \and
Korea Advanced Institute of Science and Technology (KAIST)
}

\maketitle

\newcommand{\refFig}[1]{Figure \ref{#1}}
\newcommand{\refEq}[1]{Equation (\ref{#1})}
\newcommand{\refAlg}[1]{Algorithm \ref{#1}}
\newcommand{\refSup}[1]{Supplementary \ref{#1}}
\newcommand{\refSec}[1]{Section \ref{#1}}
\newcommand{\refTab}[1]{Table \ref{#1}}
\newcommand{\refApp}[1]{Appendix \ref{#1}}

\newcommand{\Skip}[1] {
}

\newcommand{\huo}[1] {
	\textcolor{orange}{\bfseries{HUO: {#1}}}
}
\newcommand{\xie}[1] {
	\textcolor{cyan}{\bfseries{XIE: {#1}}}
}

\input{sections/abstract}

\input{sections/intro}

\input{sections/related}

\input{sections/preliminaries}

\input{sections/method}

\input{sections/training}

\input{sections/experiments}

\input{sections/conclusion}

%
%
\bibliographystyle{splncs04}
\bibliography{main}
\end{document}

%% file: sections/abstract.tex
\begin{abstract}
In this work, we use multi-view aerial images to reconstruct the geometry, lighting, and material of facades using neural signed distance fields (SDFs). Without the requirement of complex equipment, our method only takes simple RGB images captured by a drone as inputs to enable physically based and photorealistic novel-view rendering, relighting, and editing. However, a real-world facade usually has complex appearances ranging from diffuse rocks with subtle details to large-area glass windows with specular reflections, making it hard to attend to everything. As a result, previous methods can preserve the geometry details but fail to reconstruct smooth glass windows or verse vise. In order to address this challenge, we introduce a semantic regularization approach based on zero-shot segmentation techniques to improve material consistency, a frequency-aware geometry regularization to balance surface smoothness and details on different surfaces, and an analytical daylight model to enable efficient modeling of the lighting in large-scale outdoor scenes. In addition, we capture a real-world facade aerial 3D scanning image set and corresponding point clouds for training and benchmarking. The experiment demonstrates the superior quality of our method on facade holistic inverse rendering, novel view synthesis, and scene editing compared to state-of-the-art baselines. 
\end{abstract}

%% file: sections/intro.tex
\section{Introduction}


Accurately reconstructing the geometry and material for facades in outdoor scenes is a long-standing and challenging task, which could enable applications such as controllable relighting for photographs, augmented reality, digital twins, and the generation of metaverse scenes. Nonetheless, most scanning schemes like Lidar cannot reconstruct 3D models of objects as large as buildings or capture roofs. Aerial 3D scanning~\cite{roberts2017submodular} is a potential solution that captures a sequence of images of a facade using a done.  However, given multi-view images under unknown illumination, the traditional method based on oblique photography suffers from broken geometry and can't recover the material properties of outdoor buildings. Owing to the rapid development of neural implicit representation~\cite{mildenhall2020nerf}, recent works~\cite{boss2021nerd,zhang2021nerfactor,rudnev2022nerfosr} have demonstrated impressive results in reconstructing the underlying shape and material properties of objects or scenes. TensorIR~\cite{Jin2023TensoIR} proposes a novel inverse rendering approach employing tensor factorization and neural fields, which can efficiently reconstruct the shape and the material properties, i.e., albedo and roughness. However, the TensorIR is not adapted for large-scale scenes with backgrounds and complex spatially-varying materials. NeRF-OSR\cite{rudnev2022nerfosr} aims to solve the outdoor scene relighting based on neural radiance fields, while the geometry recovered by NeRF-OSR is noisy because of the inherent flaws in the NeRF-based methods. 

\Skip{
We present a 3D aerial 3D scanning dataset with images captured by a drone and ground-truth point clouds scanned by high-fidelity Lidar from a cherry picker. Equipped with those aerial images, we design a neural rendering pipeline that enables the holistic inverse rendering of facades from aerial images, providing high-quality geometry and material for novel view synthesis, relighting, and editing for a variety of downstream applications.
}
\begin{figure}[t] 
\centering 
\includegraphics[width=\linewidth]{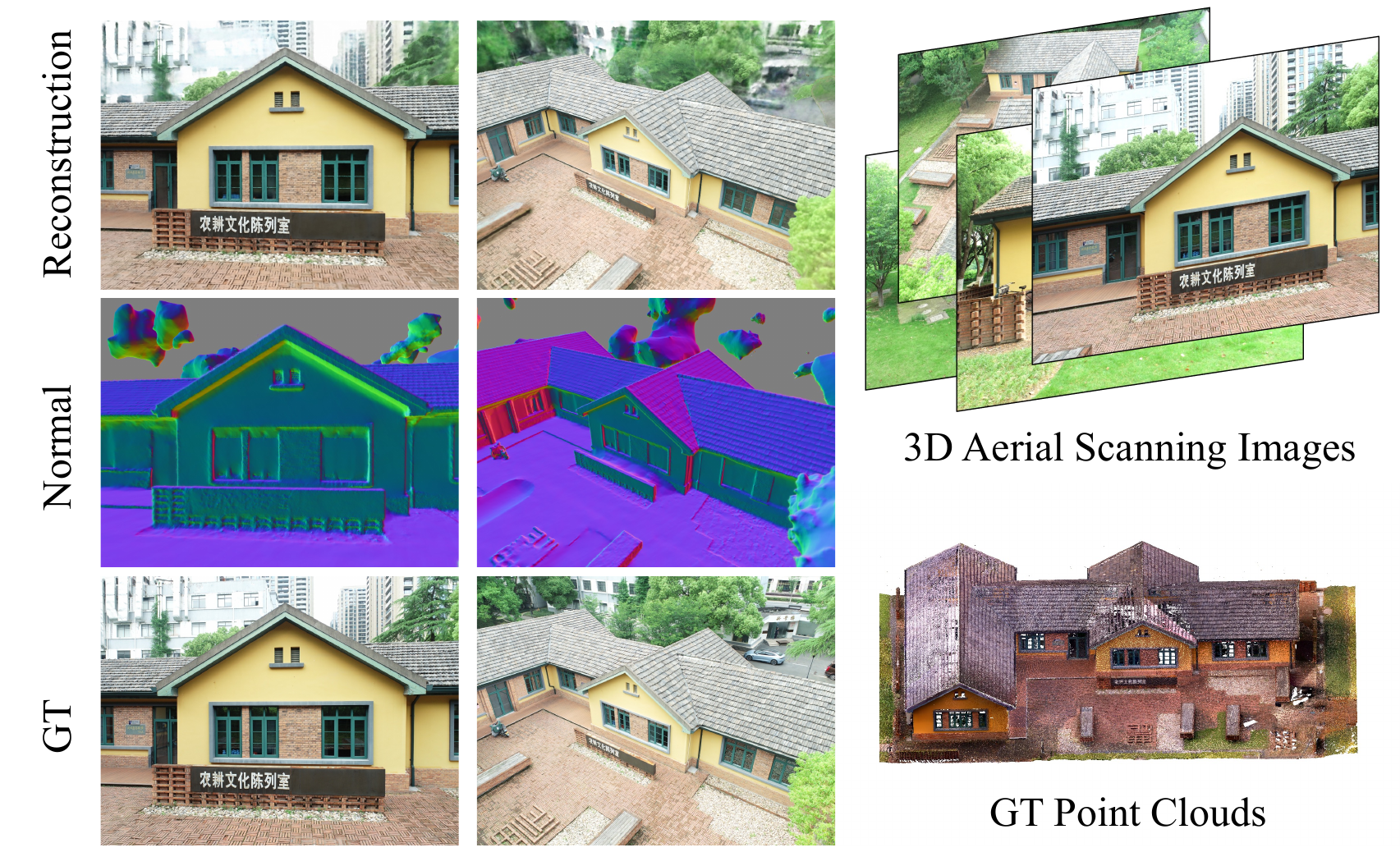}
\caption{We design a neural rendering pipeline that enables the holistic inverse rendering of facades from aerial images, providing high-quality geometry and material for novel view synthesis, relighting, and editing for downstream applications.}
\label{fig:teaser} 
\end{figure}

In this work, we present a novel holistic inverse rendering framework that enables reconstructing lighting, geometry, and material properties for the facade of the outdoor scene from multi-view aerial images. We represent the scenes as neural implicit signed distance with multi-resolution feature grids and model the diffuse color and specular color of appearance separately, and then the geometry of the scenes is optimized end-to-end using volumetric rendering with a frequency-aware smoothness constraint. To further decompose the material properties of any points in the scene surface, we parameterize the spatially-varying material properties of the scene as a neural field and introduce a semantic-adaptive material segmentation and cluster approach to regularize it. Furthermore, we represent the environment light with an analytic daylight model for full spectral sky-dome radiance. Finally, we jointly optimize the lighting and material with a differentiable Monte Carlo render layer to produce photorealistic re-rendering results. We demonstrate that our approach can reconstruct high-quality and smooth geometry as well as plausible material properties in a new aerial 3D scanning facade dataset, whose ground-truth (GT) geometry is captured by a high-fidelity Lidar from a cherry picker.

Concretely, our contributions include: 
\begin{itemize}
    \item We present an SDF-based facade reconstruction framework using aerial 3D scanning images, enabling high-quality geometry, material, and lighting reconstruction of facades containing complex materials and geometry. 
    \item We introduce a loss that leverages the latest zero-shot semantic segmentation techniques to regularize the semantic consistency of the recovered material.
    \item We design a frequency-aware geometry regularization to adaptively balance the surface smoothness and details on a wide range of materials, producing intact windows while preventing over-smoothness in other areas.
    \item We introduce the analytical daylight model to efficiently model the lighting of large-scale outdoor scenes.
    \item We contribute a facade dataset for training and benchmarking obtained by real-world aerial 3D scanning. The dataset contains multiview images captured by a drone and the corresponding Lidar-captured GT point clouds.
\end{itemize}

%% file: sections/related.tex
\section{Related Work}

\paragraph{Implicit neural scene representations.} Neural representations have become a rapidly progressing area of research to solve the problems of novel view synthesis and scene reconstruction. Neural radiance field (NeRF)~\cite{mildenhall2020nerf} uses an MLP to implicitly encode a scene into a volumetric field of density and color, and takes advantage of volume rendering to achieve impressive novel view synthesis results. Alternately, VolSDF~\cite{yariv2021volume} and NeuS~\cite{wang2021neus} implicitly represent scene geometry as a signed distance function (SDF) via a single MLP, recovering smoother surface geometries than density-based methods. Follow-up works accelerate and enhance NeRF's computationally costly MLP by hybrid representations, including dense voxel grids~\cite{yu2021plenoxels,sun2021direct},  tensors~\cite{Chen2022ECCV} or hashgrids~\cite{muller2022instant}, achieving faster training speed and better reconstruction on high-frequency details. In this paper, we adopt one of the SOTA implicit SDF method, VolSDF~\cite{yariv2021volume}, as our neural representation backbone, and leverage multi-resolution hashgrids from Instant-NGP~\cite{muller2022instant} to accelerate our representation.

\paragraph{Neural inverse rendering and relighting.} 
Inverse rendering \cite{sato1997object} is a longstanding problem that aims to decompose the lighting condition and intrinsic properties of objects in a collection of observed images. \Skip{The disentangled material and illumination enable editing and relighting applications.} Many learning-based works \cite{liu2019soft,boss2021nerd,zhang2021nerfactor} have been proposed to estimate objects' underlying shape and reflectance properties. Liu et al. \cite{liu2019soft} and Kato et al.\cite{kato2018neural} propose their novel formulation of the rasterization pipeline resulting in differentiable renders, and represent the object as a sphere template mesh that will be deformed into the shape of the object by end-to-end training. However, limited to the fixed topology of template mesh, these methods are unsuitable for reconstructing complex outdoor scenarios. Recently, a series of works \cite{boss2021nerd,kuang2022neroic,srinivasan2021nerv} focus on reconstructing implicit neural fields of objects and scenes, which accomplish great success because of the efficient representation of the implicit neural fields. \Skip{TensoIR~\cite{Jin2023TensoIR} leverages efficient tensorial representations to achieve state-of-the-art quality in single object inverse rendering. NVDIFFREC~\cite{Munkberg_2022_CVPR} designs a differentiable rasterization pipeline to jointly optimize mesh topology and underlying materials.} Recent works also succeed in scene-level inverse rendering, including indoor scenes~\cite{zhu2022learning,li2022physically,fipt2023,zhu2023i2sdf} and outdoor scenes~\cite{rudnev2022nerfosr}. NeRF-OSR~\cite{rudnev2022nerfosr} is one of the most relevant works to ours, which attempts to relight outdoor scene scenarios and introduce an outdoor dataset containing multi-view images of several buildings. However, their dataset was captured by cameras from the ground without the view covering the building roofs and had no geometry GT. In this work, we propose a new outdoor dataset that is captured using a drone from multi-view covering the whole building and provides the reference geometry from the Lidar scan.

\paragraph{Priors into neural scene representations.} Learning implicit neural representations from RGB images only may suffer from ambiguities and local minima. Recent researchers have proposed to incorporate additional priors to better supervise the scene representation. Manhattan-SDF~\cite{guo2022manhattan} adopt semantic priors from Manhattan world assumptions, while MonoSDF~\cite{Yu2022MonoSDF} exploit monocular depth and normal predictions to supervise the reconstruction of scene geometry. FIPT~\cite{fipt2023} leverage semantic priors to improve the material consistency in inverse rendering. NeRFactor~\cite{zhang2021nerfactor} introduces a data-driven prior on real-world BRDFs into the prediction network, but the generality of the prediction network depends heavily on the training data. Recently, the Segment Anything model (SAM)~\cite{kirillov2023segany} has achieved great success in zero-shot semantic segmentation, which can provide rich semantic priors for our reconstruction and inverse rendering tasks.

%% file: sections/preliminaries.tex
\section{Preliminaries}
In Neural Radiance Field (NeRF)~\cite{mildenhall2020nerf}, a scene is represented as a volumetric field of particles that emit radiance. For arbitrary position $x$, NeRF employs two MLP functions to predict the volumetric density $\sigma$ and color $c$, respectively. To generate an image from viewpoint $o$, NeRF casts ray $r$ along the direction $d$ of each pixel, and samples a set of points $x_i=o+t_id$ along the ray. The color and density of the sampled points will be integrated into pixel color using volumetric rendering as follows:

\begin{equation}
\begin{aligned}
\label{eq:volume integrate}
    C(r) = \sum_{n=1}^{N}(\prod_{m=1}^{n-1}(1-\alpha_m)\alpha_n c_n), \alpha_n = 1-exp(-\sigma_n\delta_n),
\end{aligned}
\end{equation}
where $\delta_n = t_{n+1}-t_{n}$ is the interval between sample $n$ and $n + 1$, $c_n$ and $\sigma_n$ are the color and density of sampled points, respectively. $\alpha_n$ represents the transmittance of the ray segment between sample points $x_{i-1}$ and $x_i$. 

With input views, a set of observed images with calibrated camera information, NeRF optimizes the MLP parameters by minimizing the L2 difference $\mathcal{L}_{col}$ between the reference pixel color from the input image and the corresponding pixel color predicted using volumetric rendering, where $r\in R$ and $R$ denotes a set of sampled rays. In order to reconstruct the large unbounded scenes with 360 captures, we follow NeRF++\cite{2010.07492} to partition the scene space into two volumes, an inner unit sphere and an outer volume corresponding to foreground and background, respectively. We simply reconstruct the background portion as a NeRF field. However, for the foreground part, we use implicit SDF to formulate object geometry and aim for exceptional geometric quality. Inspired by Instant NGP \cite{muller2022instant}, we introduce multi-resolution feature grids $\left\{\Phi_\theta^l\right\}_{l=1}^L$ of resolution $R_l$ into our implicit representation to encode spatial position $x$ with multiple features. Given a 3D point $x$, we employ the geometry network $F_g$ maps it to the SDF value $s$ and feature $z$:

\begin{equation}
 (s,z, \rho, q)=F_g\left(\gamma(x),\left\{\operatorname{interp}\left(x, \Phi_\theta^l\right)\right\}_l\right) ,
\label{eq:geometry}
\end{equation}
where $\rho$, $q$ is the specular hint of the point $x$, and these factors will be utilized for color estimation as mentioned in \refEq{eq:specular_color}. In addition, interp is the trilinear interpolation, and $\gamma(x)$ corresponds to frequency encodings introduced by \cite{vaswani2017attention}. Following VolSDF~\cite{yariv2021volume}, we transform the SDF value $s$ to volume density $\sigma$ using the cumulative distribution function of the Laplace distribution. For more details, please refer to the appendix.

\Skip{
\begin{equation}
\mathcal{L}_{col}=\sum_{r\in R}\left \| C_{gt}(r)-C(r)\right \|_{2},
\label{eq:colorl1loss}
\end{equation}
where $R$ denotes a set of sampled rays.
}

\Skip{
In order to reconstruct the large unbounded scenes with 360 captures, NeRF++\cite{2010.07492} proposes to partition the scene space into two volumes, an inner unit sphere and an outer volume corresponding to foreground and background, respectively. For 3D point $(x,y,z)$ out of the unit sphere, NeRF++ proposes a technique named inverted sphere parameterization to re-parameterize the point by a quadruple $(x', y',z', 1/\text{r})$, where $x'^2 + y'^2 + z'^2 = 1$, $\text{r}=\sqrt{x^2+y^2+z^2}$, and $1/\text{r}$ means the inverse distance. The points sampled from the two volumes are fed to two MLPs for predicting the density and color of points separately. As a result, the final color along a ray $r$ can be calculated as the sum of foreground color $C_{fg}$ and  background color $C_{bg}$ :
\begin{equation}
\begin{aligned}
    C(r) = C_{fg}(r) +  T_{fg}* C_{bg}(r), T_{fg}=\prod_{m=1}^{N}(1-\alpha_m),
\end{aligned}
\end{equation}
where $T_{fg}$ is transmittance though the $N$ sampled points from foreground. Both $C_{bg}$ and $C_{fg}$ are calculated using \refEq{eq:volume integrate}.
}

%% file: sections/method.tex
\section{Method Overview}
\begin{figure*}[!htp] 
\centering 
\includegraphics[width=\linewidth]{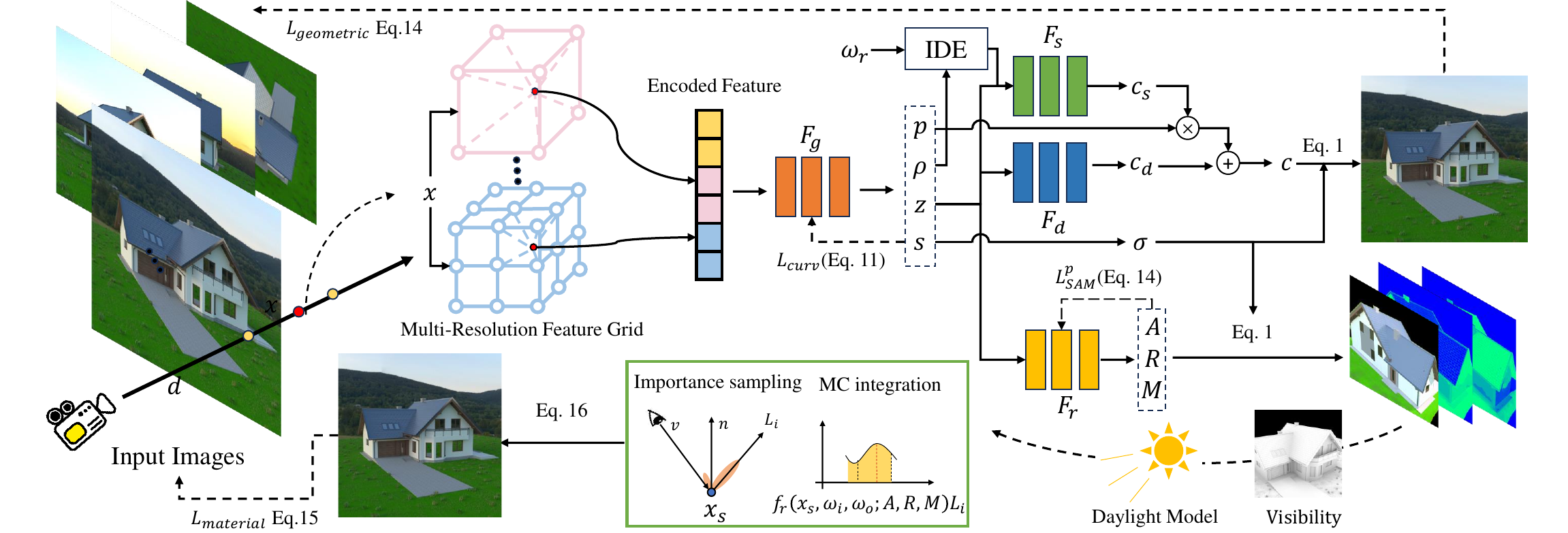}
\caption{Overview of our framework. Our method takes as input aerial multi-view images and reconstructs the full 3d facade containing geometry and material properties in two stages (two dotted lines). In the first stage, using volumetric rendering, we optimize specular color field $F_s$, diffuse color field $F_c$, and geometry network $F_s$ end-to-end. In the second stage, we decompose light and material by jointly optimizing the analytic daylight model and material field with a differentiable Monte Carlo render layer.}
\label{fig: overview} 
\end{figure*}

\refFig{fig: overview} shows the overview of our methods. We propose a novel framework that takes images of outdoor scenes captured by drones as input to holistically reconstruct the geometry, lighting, and material of the facade, which enables relighting outdoor scenes in arbitrary novel views and light conditions. Specifically, we divide this task into two stages.


In the first stage, we aim to reconstruct high-quality geometry and appearance of the complex outdoor scenes using differentiable volume rendering. We represent the scene as an implicit signed distance function (SDF) and introduce multi-resolution feature grids to encode the spatial position. To alleviate the ambiguity caused by glass reflection, we introduce $F_d$ and $F_s$ to predict specular color and diffuse color, respectively. In addition, we optimize the SDF field of the scene under the condition of curvature loss and Eikonal loss.

In the second stage, we focus on material decomposition. To solve the ill-conditioned inverse problem, we represent the spatially-varying material of the scene as a neural field $F_r$ and propose an adaptive material segmentation and cluster approach to condition the material estimation. Furthermore, we represent the environmental light using an analytical daylight model. They will be jointly optimized with the learnable material field $F_r$ employing a physically-based rendering pipeline.

\section{Neural Scene Reconstruction}
We represent the outdoor scene with the neural implicit SDF and color fields and aim to reconstruct both the geometry and appearance in high quality.


\Skip{
We anneal $\beta$ according to the strategy introduced by BakedSDF~\cite{yariv2023bakedsdf}:
\begin{equation}
 \beta_{t}=\beta_{0}(1+\frac{\beta_0-\beta_1}{\beta_0} t^{0.8})^{-1},
\end{equation}
where $t$ goes from 0 to 1 during training, $\beta_{0}$ and $\beta_{1}$ is 0.1 and 0.001 respectively.
}

\subsection{Modeling View-Dependent Appearance}
\label{sec:separate_color}

Because the glasses on building facades often have severe reflections, directly optimizing the shape of those scenes with a typical design of NeRF~\cite{mildenhall2020nerf} would produce broken glass surfaces. The reason is that the surface predicted by SDF lies between the real depth of the reflective surface and the virtual depth of the reflected image~\cite{Guo_2022_CVPR}. To tackle this problem, we separate the specular color and diffuse color of the outgoing radiance. Then a specular color field $F_s$ and a diffuse color field $F_d$ are employed to predict the two components respectively. The diffuse color field $F_d$ takes as input feature vector $z$ and point $x$ and predicts the diffuse color of point $x$, which is independent of the observed direction:
\begin{equation}
 c_d =F_d(x, z).
\label{eq:diffuse_color}
\end{equation}
Furthermore, the specular color field is expected to capture the view-dependent appearance caused by the reflected radiance of the smooth components of facades such as flat glass. Following Ref-NeRF~\cite{verbin2022refnerf}, we feed the distribution of reflection vectors IDE$(\hat{\omega}_r,\kappa)$ to specular color field $F_s$ instead of a single vector, where $\kappa$ is a concentration parameter defined as inverse roughness $\kappa = \frac{1}{\rho}$ and  Integrated Directional Encoding (IDE) is a encode technique that encodes the distribution of reflection directions using the expected value of a set of spherical harmonics (see supplementary for more details). Thus the specular color field $F_s$ is defined as:

\begin{equation}
 c_s =F_s(x, z, \rho, \operatorname{IDE}(\hat{\omega}_{r},\kappa), n\cdot d, q),
\label{eq:specular_color}
\end{equation}
where $d$, $\hat{\omega}_{r}$, $n$ are the view direction, reflection vector, and normal of point $x$, respectively. In practice, $n = \frac{\nabla_x(F_g(x))}{\|\nabla_x(F_g(x)) \|}$  is calculated as the gradient of the signed distance $s$ at point $x$. Finally, we get the color value of the point $x$ by combining the diffuse color $c_d$ with specular color $c_s$ according to specular hint $q$ as:
\begin{equation}
 c =c_d+q\odot c_s ,
\label{eq:final_color}
\end{equation}
where $\odot$ denotes elementwise multiplication.

\subsection{Frequency-aware SDF Regularization}
\label{sec:sdf_regular}

High-frequency reflected light from the glass of the building facades often creates ambiguity in geometry optimization, leading to an uneven glass surface in the reconstructed scene. In order to reconstruct a flat glass surface, we apply an approximate curvature loss~\cite{rosu2023permutosdf} on SDF. Given a 3d point $x$, we compute a tangent vector $\eta$ by the cross product of its normal $ n$ with a random unit vector. Then we obtain a new perturbed point $x_\epsilon$ along the tangent vector $\eta$, as well as its normal $n_\epsilon$. The approximate curvature loss defines as:
\begin{equation}
 \mathcal{L}_{curv}=\sum_x w_i\cdot(n * n_\epsilon -1)^2,  w_i=lap(\kappa),
\label{eq:curvature_loss}
\end{equation}
where $lap$ is a Laplace function with the $\mu$=1 and $b$=1, which enable us to apply curvature loss adaptively according the specular factor of different surface areas.
In addition, we use an Eikonal loss for any point x to regularize SDF:
\begin{equation}
 \mathcal{L}_{eik}=\sum_{x}( \| \nabla_x(F_g(x)) \|-1)^2.
\label{eq:eikonal_loss}
\end{equation}

\section{DECOMPOSITION VIA INVERSE RENDERING}
We detail the method for material formulation and semantic segmentation, and introduce the analytical daylight model and differentiable Monte Carlo layer.

\subsection{Material Formulation and Semantic Segmentation}
\label{sec:semantice_seg}

We parameterize the spatially-varying material of the scene as a neural field $F_r$, which takes as input the point $x$ and feature vector $z$ to estimate the albedo $A$, roughness $R$, and metallic $M$ for any point. 
\begin{equation}
 (A, R, M) =F_r(x, z).
\label{eq:brdf_net}
\end{equation}
While $F_r$ could be jointly optimized with the physically-based renderer, the inferred material property suffers severe artifacts because of the ill-conditioned nature of material decomposition. To address this problem, we propose an adaptive material segmentation and cluster approach to condition the material estimation.

\begin{wrapfigure}{r}{0.5\textwidth}
    \centering
    \includegraphics[width=1\linewidth]{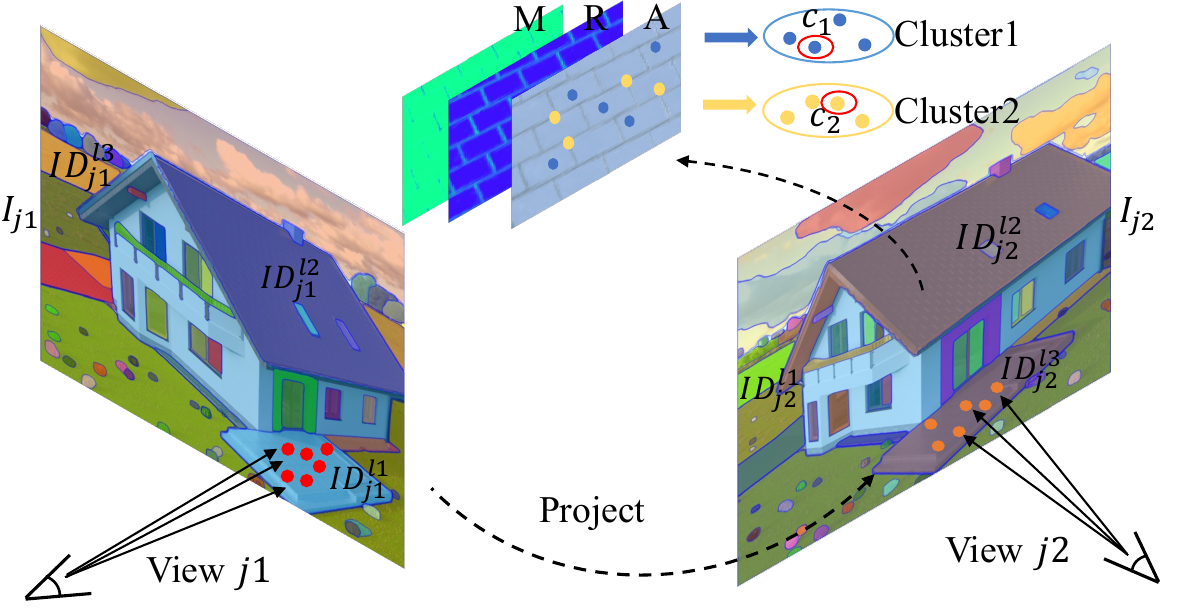}
    \caption{Illustration of the SAM loss. A 2D semantic instance is projected to other views to match its correspondence instance, and then the material properties are regularized to the cluster centers in the same instance.}
    \label{fig: segment}
\end{wrapfigure}

 Segment Anything Model (SAM)~\cite{kirillov2023segany} has demonstrated its ability to segment any instance of an image. For training data containing $M$ input images, we apply SAM to all input images  $\left\{I_1, I_2...I_j, j\in M \right\}$ to segment each image to instances as well as the corresponding segmentation id $ID_j^{l}$ of each instance, where $ID_j^{l}$ means the instance in image $j$ numbered $l$. As the SAM only performs on the 2D region of the input image, the same instance has different segmentation ids in multi-view images, e.g., as shown in \refFig{fig: segment}, the step is assigned $l_1$ in view $j_1$ while assigned $l_3$ in view $j_2$. We need a strategy to match the different ids from multiple views to ensure an instance has a unique id in all input images. Benefiting from the full 3D sense representation reconstructed by volume rendering, we can project the pixels sampled in an image $I_{j1}$ to the adjacent image $I_{j2}$. Pixels belonging to an instance in image $I_{j1}$ may cover different segmentations in image $I_{j2}$ because of the inaccurately estimated camera poses in real data. We use the Hungarian matching algorithm \cite{kuhn1955hungarian} to tackle this problem to match the instance ids in multiple views.

Equipped with the instance segmentation result, we introduce an adaptive smoothing loss to 
constrain the material properties of each instance during inverse rendering. Specifically, we sample a set of $M$ pixels within the region of segmentation for each instance and perform Kmeans~\cite{hartigan1979algorithm} clustering on its albedo $A$, roughness $R$, and Metalic $M$, respectively. As a result, the sampled pixels are divided into $K$ clusters, and the pixels in the same cluster are considered to have similar material properties. Hence, we adopt a Hungarian loss  as SAM loss to smooth the properties of those pixels:
\begin{equation}
 \mathcal{L}_{SAM}^{p}=\sum_{i}^M\|  p_i - \mathop{\arg\min}\limits_{c_k}\|p_i-c_k\| \|_1,
\label{eq:hungarian_loss}
\end{equation}

where $p_i$ is the property value of pixels within the current instance segmentation, $c_k$ is the $k$-th cluster center. This smoothness constraint helps to optimize the material properties, e.g., remove artifacts caused by shadow and geometry changes from albedo. To maintain texture variety and detail while smoothing the material, the number $K$ of clusters will be dynamically updated for the optimized material properties. We cluster the pixels into k-1, k, and k+1 clusters, respectively, and calculate the silhouette coefficients \cite{rousseeuw1987silhouettes} of the clusters. Then, the new cluster number $K$ will be updated to the scheme, which leads to the minimum silhouette coefficients.

\subsection{Analytical Daylight Model}
\label{sec:sunlight}
We discovered that directly optimizing a skybox texture as the incident light makes it challenging to recover the material properties of a building's facade. This is because the skybox texture has an extensive parameter space, potentially leading to the optimization of unrealistic skies. As the images of an outdoor scene are captured under daylight conditions, we introduce an analytical sunlight model~\cite{sundaymodel}, allowing for the analytical generation of the sky's lighting distribution with only minimal optimization parameters. The incident radiance from direction $\omega_i$ are calculated as:
\begin{equation}
\begin{aligned}
    \mathbb{F}(\theta,\gamma)= (1+Ae^{\frac B{\cos\theta+0.01}})\cdot(C+De^{E\gamma} + F\cos^2\gamma +\\
    + G\cdot\frac{1+\cos^2\gamma}{(1+H^2-2H\cdot\cos\gamma)^{\frac32}}+I\cdot\cos^{\frac12}\theta) 
\end{aligned}
\label{eq:sunlight}
\end{equation}
where $\theta,\phi$ is the azimuth angle of the sun, which determines the solar position and $\gamma$ is the angle between the incident direction and the solar point. We adhered to the implementation in Mitsuba~\cite{Mitsuba} and developed a differentiable version, incorporating the solar position and turbidity as optimizable parameters.
\begin{wrapfigure}{r}{0.5\textwidth}
    \centering
    \includegraphics[width=1\linewidth]{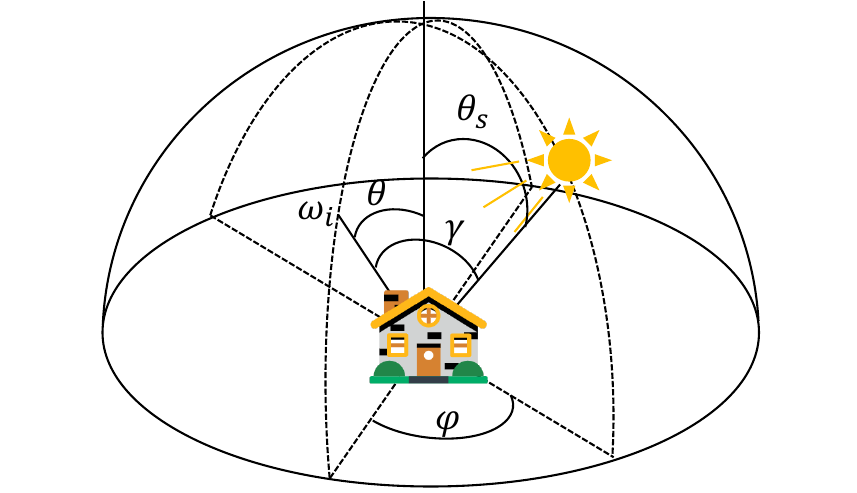}
    \caption{The incident radiance from the sun received by the building from any incident direction $\omega_i$. More discussion in ~\refSec{sec:sunlight}}
    \label{fig: sunday}
\end{wrapfigure}
The $A,B,C,D,E,F,G,I$ are parameters that could be calculated as a function of solar position $\theta,\phi$ and turbidity $\tau$ ~\cite{ineichen1994sky,sundaymodel}. Therefore, once the position of the sun and the turbidity of the sky are determined, the distribution of the sky's illumination is also established. Solar position and turbidity are the only parameters we focus on, which significantly reduces the number of parameters compared to optimizing a skybox texture. The resulting distribution of sunlight is also much more consistent with real-world conditions. For more details on the analytical daylight model, please refer to the appendix.

\subsection{Physically-based Rendering}
\label{sec:pbr_render}

To decompose the facade into underlying geometry and material properties, we introduce a differential Monte Carlo raytracing to produce photorealistic re-rendering results. We calculate the outgoing radiance $L_o$ in direction $\omega_o$ follow the general rendering equation~\cite{kajiya1986rendering}: 
\begin{equation}
    L_o(x_{s}, \omega_o) = \int_\Omega f_r(x_{s}, \omega_i, \omega_o)L_i(x_{s}, \omega_i)(\omega_i \cdot n_s))d\omega_i,
\label{eq:renderingeq}
\end{equation}
where $\omega_i$ is the incident direction, $\omega_o$ is the outgoing direction, $x_{s}$ is a surface point,  $f_r(x_{s}, \omega_i, \omega_o)$ is the BRDF term, $L_i(x_{s}, \omega_i)$ is the incident radiance from direction $\omega_i$, and the integration domain is the hemisphere $\Omega$ around the surface normal $n_s$ of the intersection point, which is calculated using ~\refEq{eq:volume integrate}. Regarding the BRDF term, we parameterize it with the popular Cook-Torrance analytical BRDF model with the GGX ~\cite{walter2007microfacet} normal distribution function (NDF) and denote it $f_r(x_{s}, \omega_i, \omega_o; A, R, M)$. Note that, the $A$, $R$, $M$ here are albedo, roughness, and metallic value of surface point $x_s$, which could be computed using ~\refEq{eq:volume integrate}. 

Then, we could estimate the light radiance from the analytical daylight model, which is jointly optimized with the material field $F_r$.
In our work, we consider second-bounce reflection, which requires us to determine whether the reflected light intersects the scene or hits the background. Obviously, we could calculate the opacity for arbitrary direction from any surface point of the scene using ray marching. However, this approach becomes prohibitively expensive when increasing the sample rays. To alleviate this issue, we utilized the Marching Cubes~\cite{10.1145/37402.37422} algorithm to extract mesh geometries, through which we could compute intersection rapidly.

In summary, our render layer computes the unbiased image as:
\begin{equation}
\label{eq:is_estimator}
    \hat{L}_{o}(x_{s},\omega_o)= \frac{1}{m}\sum_{i=1}^{m}\frac{\Gamma_{sg}(\omega_i)f_r(x_{s}, \omega_i, \omega_o; A, R, M)(\omega_i \cdot n)}{p(x_{s}, \omega_i, \omega_o; A, R, M)},
\end{equation}
where $p(x_{s}, \omega_i, \omega_o; A, R, M)$ is the probability distribution function (PDF) for importance sampling. $m$ is the number of the ray directions sampled on the point $x_s$ using the importance sampling.

%% file: sections/training.tex
\section{Training}

Our training in outdoor scene reconstruction and intrinsic decomposition are conducted in two stages: 
In the first stage,  training geometry and color fields with the loss:
\begin{equation}
    \mathcal{L}_{\mathrm{geometric}} = \mathcal{L}_{\mathrm{col}}+\lambda_{\mathrm{1}}\mathcal{L}_{\mathrm{curv}}
    +\lambda_{\mathrm{2}}\mathcal{L}_{\mathrm{eik}},
\end{equation}
where $\lambda_1$ is initialized as 1 and will gradually decrease to 0, $\lambda_2$ is 0.1.
In the second stage, training the material field, environment light, and volumetric visibility grid with the loss:
\begin{equation}
\begin{split}
    \mathcal{L}_{\mathrm{material}} = \mathcal{L}_{\mathrm{col}}+
    \lambda_{\mathrm{1}}\mathcal{L}_{SAM}^{\mathrm{albedo}}+ 
    \lambda_{\mathrm{2}}\mathcal{L}_{SAM}^{\mathrm{roughness}}+
    \lambda_{\mathrm{3}}\mathcal{L}_{SAM}^{\mathrm{metallic}},
\end{split}
\end{equation}
where $\lambda_1,\lambda_2,\lambda_3$ are all set to 0.1, $\mathcal{L}_{\mathrm{col}}$ is the $L_1$ loss between rendering color produced by Monte Carlo render layer and reference color. For more training details, please refer to the appendix.

\Skip{
In the first stage, we initialize our network with the SDF of a sphere presented in \citet{atzmon2020sal}. Then, we optimize the geometry network $F_g$ and color fields $F_s,F_d$ using the following loss:
\begin{equation}
    \mathcal{L}_{\mathrm{geometric}} = \mathcal{L}_{\mathrm{col}}+\lambda_{\mathrm{1}}\mathcal{L}_{\mathrm{curv}}
    +\lambda_{\mathrm{2}}\mathcal{L}_{\mathrm{eik}},
\end{equation}
where $\lambda_1$ is initialized as 1 and will gradually decrease to 0, $\lambda_2$ is 0.1. With the observation that areas with bigger $\kappa$ value should have more smooth geometry, we apply different strategies to update the $\lambda_1$. For areas with $\kappa$ larger than the mean value in \refEq{eq:specular_color}, $\lambda_1$ will decrease to 0 in 1000 epochs, when the $\lambda_1$ will decrease to 0 in 100 epochs in the other area. $\mathcal{L}_{\mathrm{col}}$ is the $L_1$ loss between rendering color produced by volumetric rendering and reference color. In the second stage, we first train the volumetric visibility grid $\Lambda_{sg}$ with 6 spp samples in the first 100 epochs, where the reference visibility is obtained by ray marching the SDF field. After the warmup training, we begin to optimize the material field $F_r$, environment light $\Gamma_{sg}$. We increase the spp to 512 and use the trained volumetric visibility grid instead of ray marching for visibility query to accelerate the training. We add the Hungarian loss to albedo $A$, roughness $R$, and metallic $M$ to smooth the material predicted by the material field. We initialize the numbers of cluster centers with ${2,2,1}$ for roughness, metallic, and albedo, respectively. In subsequent training, the number of cluster centers will be updated every 50 epochs according to the strategy introduced in the later section. The loss of the material decomposition stage is:
\begin{equation}
\begin{split}
    \mathcal{L}_{\mathrm{material}} = \mathcal{L}_{\mathrm{col}}+
    \lambda_{\mathrm{1}}\mathcal{L}_{SAM}^{\mathrm{albedo}}+ 
    \lambda_{\mathrm{2}}\mathcal{L}_{SAM}^{\mathrm{roughness}}+
    \lambda_{\mathrm{3}}\mathcal{L}_{SAM}^{\mathrm{metallic}},
\end{split}
\end{equation}
where $\lambda_1,\lambda_2,\lambda_3$ are all set to 0.1, $\mathcal{L}_{\mathrm{col}}$ is the $L_1$ loss between rendering color produced by Monte Carlo render layer and reference color.

In addition, Adam optimizer~\cite{kingma2014adam} is employed for optimizing all networks and parameters. We set the learning rate to $2\times10^{-3}$ with an exponential falloff during the optimization. We sample 1024 ray directions each iteration. The entire training process takes approximately 15 hours with a single NVIDIA V100 GPU.
}

%% file: sections/experiments.tex
\begin{figure}[ht] 
\centering 
\includegraphics[width=\linewidth]{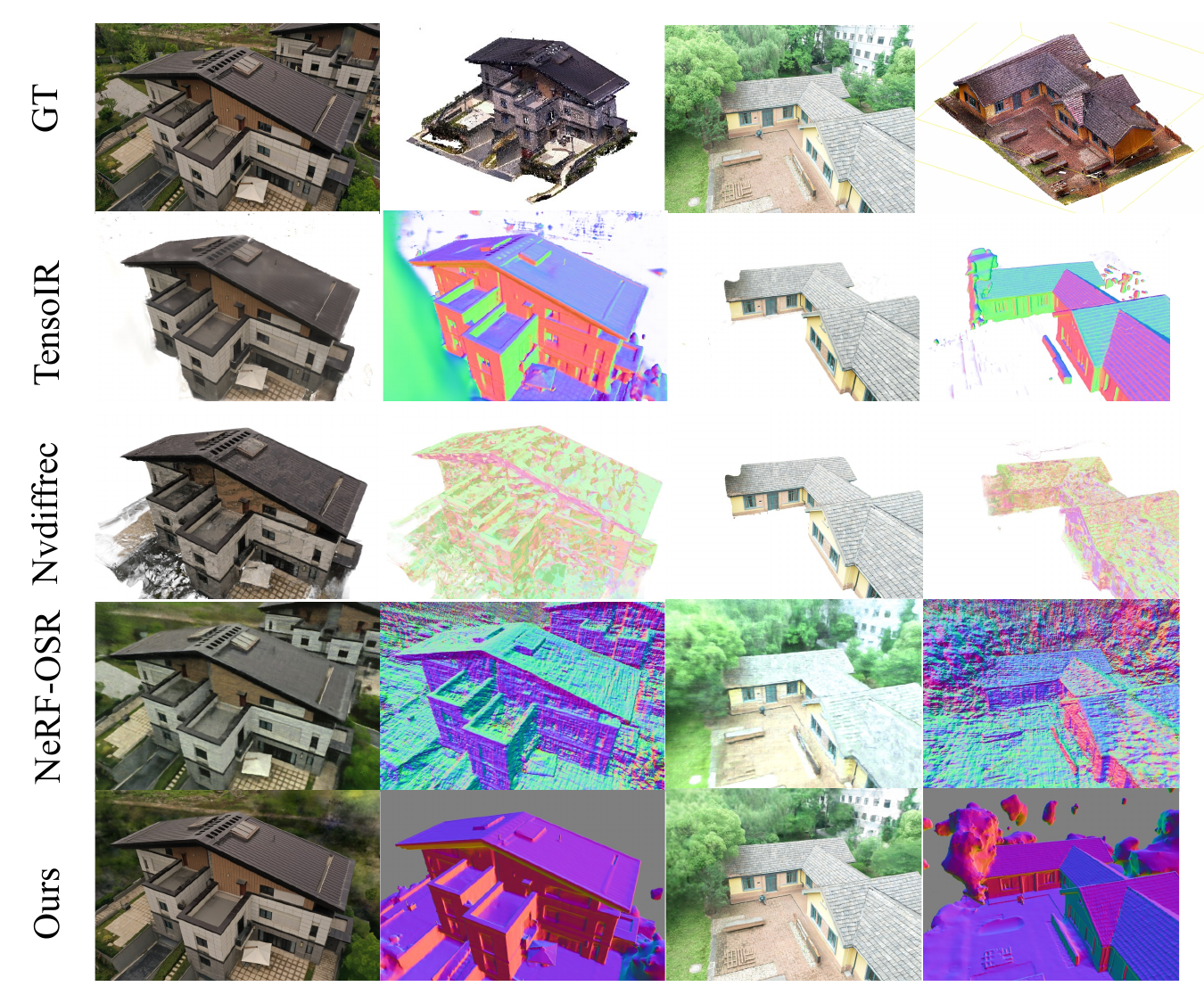}
\caption{Comparison on two real-world scenes with TensoIR, Nvdiffrec, and NERF-OSR. We visualize the novel-view results and reconstructed normals. Note that the backgrounds and the trees overlapping the buildings are not our reconstruction targets and are not covered by the scanning. }
\label{fig:real_comparison} 
\end{figure}

\begin{figure}[!htp] 
\centering 
\includegraphics[width=\linewidth]{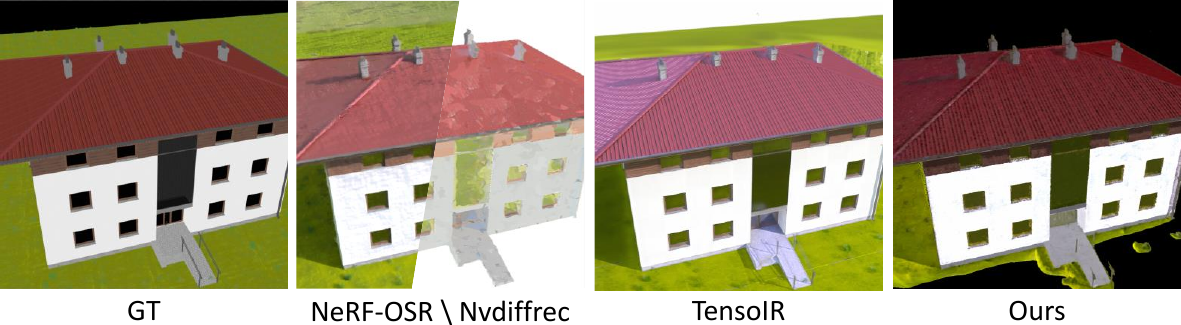}
\caption{Comparison on the synthetic dataset for the estimated albedo. Our method could produce more convincing results. For instance, our method can remove the shadow of the chimneys from the albedo and produce semantically consistent albedo. }
\label{fig:syn_comparison_material} 
\end{figure}

\section{Experiments}\label{sec:experiments}

\subsection{Experiment Settings}

\paragraph{Datasets.} We demonstrate our approach on both synthetic and self-captured real-world datasets. Specifically, the synthetic dataset is rendered by Blender using the Cycles engine, and each building has 200 randomly sampled views distributed around the hemisphere of the building. We also provide the GT camera poses for each produced image. For the real-world dataset, we operate the drone to fly around the building at different heights and take pictures at random sampling points. Each real-world dataset contains about 300 images, and we use Colmap\cite{schoenberger2016sfm} to calibrate the camera parameters of these images. For each training dataset, we select 20 views to evaluate and use the rest of the views for training. In order to evaluate the quality of the reconstructed geometry, we managed to scan the point cloud of the building with a high-fidelity Lidar~\cite{TrimbleX7}. To capture all aspects, especially the roof, of the building, we employ a cherry picker to lift the Lidar to designated positions around the building and scan the point cloud of the building from those positions. 

\paragraph{Baselines.} We compare our method with NeRF-OSR\cite{rudnev2022nerfosr}, which is the state-of-the-art (SOTA) outdoor reconstruction method. In addition, We also compare our method with TensoIR\cite{Jin2023TensoIR} and Nvdiffrec\cite{Munkberg_2022_CVPR}, which are SOTA object-oriented inverse rendering frameworks. \Skip{As Nvdiffrec and TensorIR are designed for small-scale indoor objects, their results on large-scale outdoor scenes are merely a qualitative reference. }Because Nvdiffrec and TensoIR need the background mask of GT, we use SAM to generate masks of the real scene. Note that all numerical results are measured using the same mask, i.e., the background is not calculated.

\subsection{Comparsion}

We compare our method with the SOTA outdoor method, NeRF-OSR, and SOTA object-oriented methods, Nvdiffrec and TensoIR, in real-world scenes captured by 3D aerial scanning in \refFig{fig:real_comparison}. It is easy to see that our method produces sharp texture and smooth geometry details, attributed to our grid-based neural representation and geometry loss. In contrast, the color and geometry of NeRF-OSR are blurred and broken, such as the lines on the roof of the villa scene all disappearing(the first two rows). In addition, the rendered color of NeRF-OSR is biased from the original images, which we think is a result of the biased rendering model it employs. Despite the additional mask for the scene background, TensoIR still produces light spots on the roof and geometry artifacts around the corner of the walls and the windows, showing that it is unable to handle highly specular glass material of facades. Similarly, the results of Nvdiffrec are full of artifacts in both geometry and colors, as it is not designed for outdoor scenes. Compared with all of these SOTA methods, our method reconstructs smooth and detailed geometry as visualized by the normal, enabling realistic relighting results in novel lighting conditions. However, all methods fail to reconstruct the laundry racks on the balcony since they are too thin in the captured views. It is possible to resolve this problem by adaptively planning the aerial 3D scanning path to gather more close-range views in the future.

Table \ref{tab:delit compare} \Skip{and \ref{tab:geocompare} } reports the numerical results of all methods. Our method has a comparable PSNR with the TensorIR in the synthetic dataset, and ours has the highest PSNR in the real-world dataset. We believe the difference is due to the mask for the background. While in the syntheic dataset, there exist perfect object masks, which significantly help increase the PSNR. However, in the real world, the mask is not perfect. Therefore, our background and daylight model are more suitable for real-world data. More numerical compare results for reconstructed geometry can be found \textbf{in the supplementary document}.

\input{tables/comparsion_syn}

\Skip{
\subsection{Comparsion on synthetic dataset}

\refFig{fig:syn_comparison} shows the results of our geometry reconstruction on our synthetic dataset. The first building is featured with many highly glossy glass windows, and the second building has fine geometry details. In general, all methods can generate acceptable re-rendered images. However, the geometry of NeRF-OSR and Nvdiffrec suffers from serious artifacts like holes, noises, and patterns. While these artifacts are not visible in novel-view synthesis, they are undesirable for relighting applications as the artifacts would lead to many wrong shadows and light spots. In contrast, the geometry of TensorIR and ours are much smoother. The difference is that our method successfully produces smooth glass, but TensorIR produces holes in the glass area. We believe recovering the varying components of buildings, such as windows with strong reflections, is one of the most challenging parts of facade reconstruction. The results demonstrate that our method is clearly superior to others in this task attributed to the power network backbone, geometry constraint, and semantic material regularization.
}
In addition, we compare the decomposed material parameter, albedo, in the synthetic dataset as shown in \refFig{fig:syn_comparison_material}. Thanks to the SAM loss providing additional regularization on the roof, our method can remove the shadow of the  chimneys from the albedo. In contrast, all other methods cannot distinguish whether the black tint comes from shadowing or textures, leading to artifacts in the albedo. More comparison results on the synthetic datasets and real-world datasets can be found  \textbf{in the supplementary document}.

\Skip{
\begin{figure*}[!htp] 
\centering 
\includegraphics[width=\linewidth]{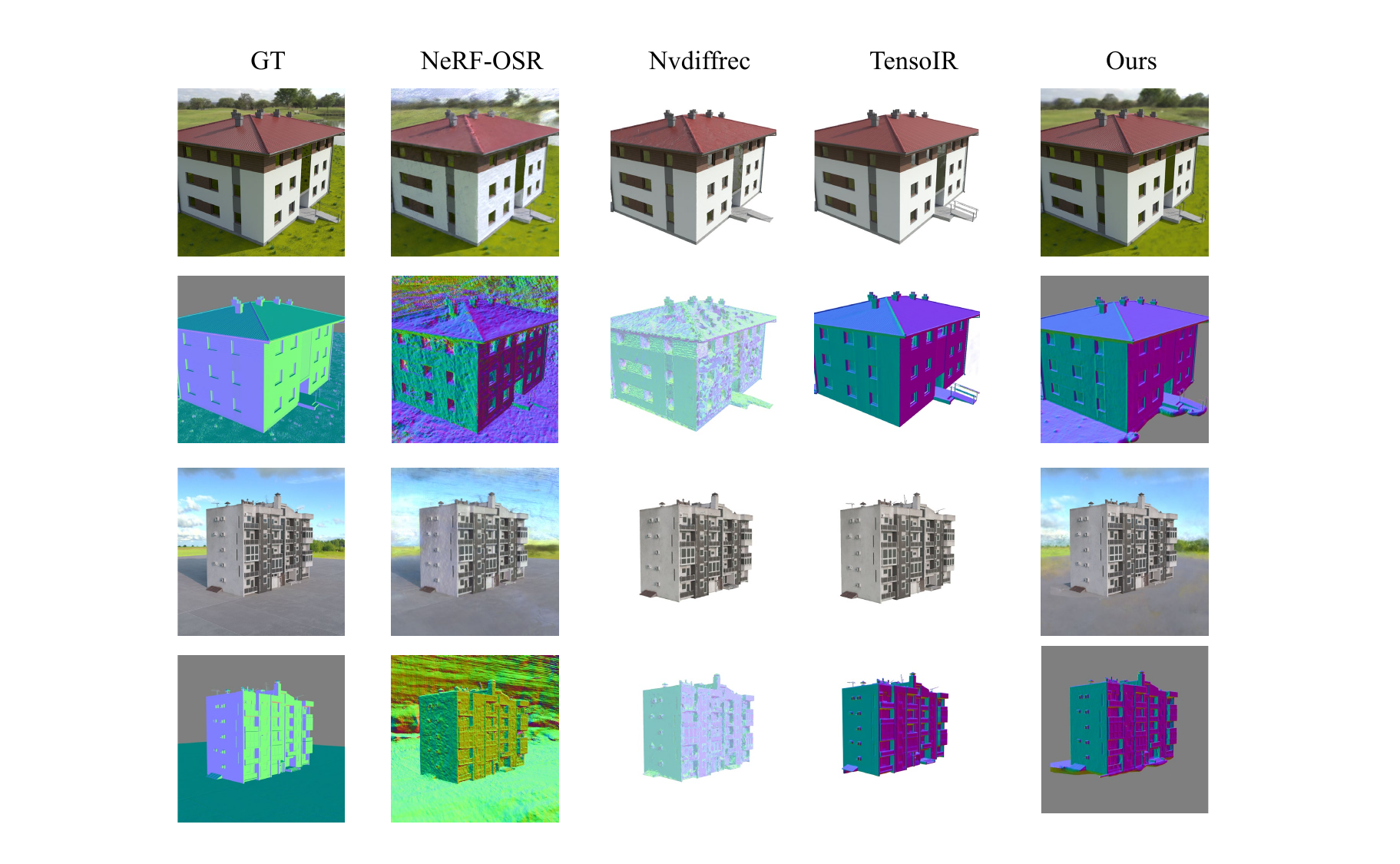}
\caption{Comparison on our synthetic dataset with NeRF-OSR, Nvdiffrec, and TensorIR. We visualize the novel-view results and reconstructed normals. Nvdiffrec and TensorIR are trained with images clipped with GT masks. NeRF-OSR and ours are trained with images with backgrounds.}
\label{fig:syn_comparison} 
\end{figure*}
}

\begin{figure}[!htp] 
\centering 
\includegraphics[width=\linewidth]{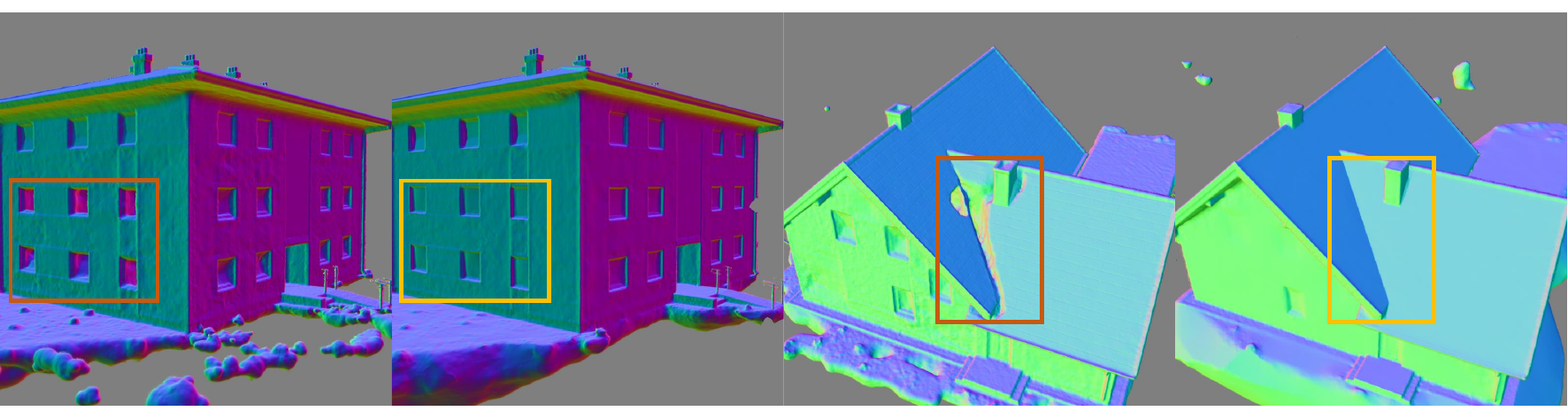}
\caption{Abalation study on the curvature loss. }
\label{fig:ablationstud1} 
\end{figure}


\subsection{Ablation Studies}

\begin{figure}[!htp] 
\centering 
\includegraphics[width=\linewidth]{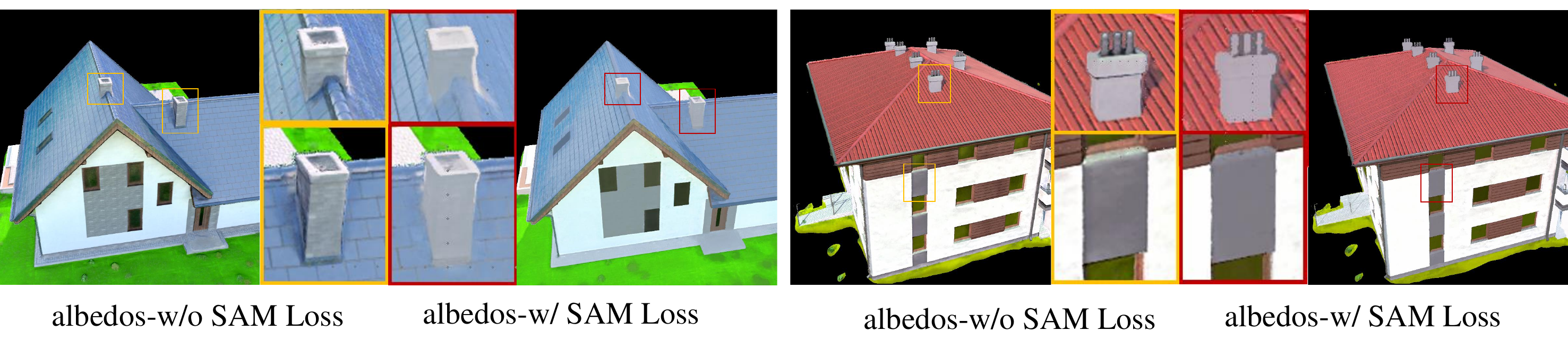}
\caption{Ablation study on the SAM loss by comparing the reconstructed albedo map. }
\label{fig:ablationstud2} 
\end{figure}

We first conduct an ablation study on curvature loss. As shown in \refFig{fig:ablationstud1}, curvature loss is the key to preventing holes in the specular windows and artifacts on geometry boundaries. Next, we test the effectiveness of the SAM loss in \refFig{fig:ablationstud2}. The results demonstrate that the SAM loss is beneficial to maintain material consistency within the same semantic instance, i.e., the roof. Without the SAM loss, some pixels wrongly regard the shadows as the albedo color of the roof, producing many noisy artifacts in the shadow areas. 

\begin{figure}[!h] 
\centering 
\includegraphics[width=\linewidth]{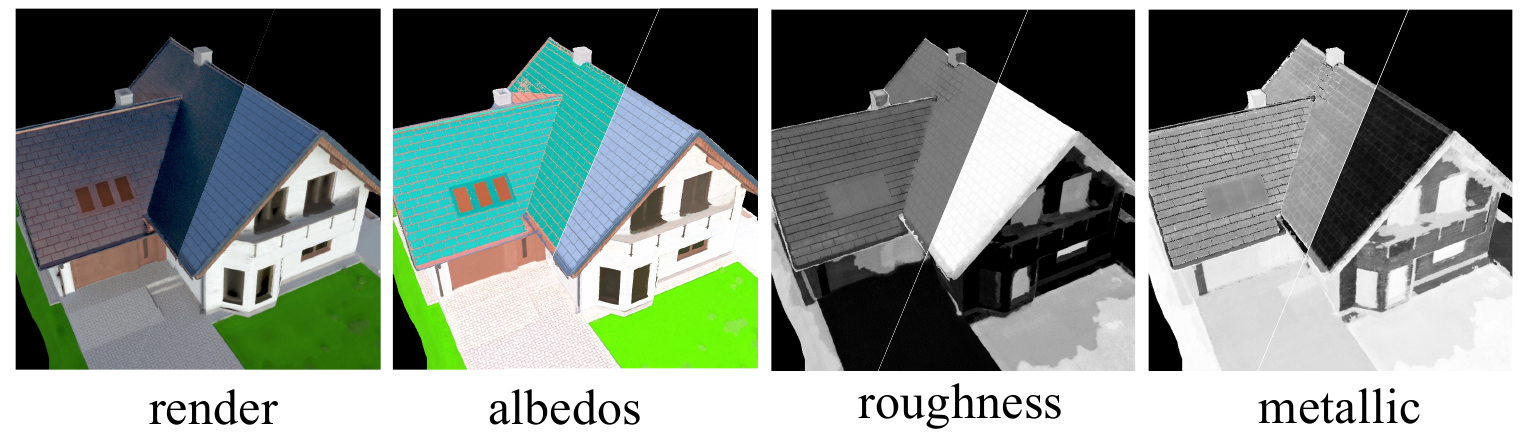}
\caption{Results of material editing. }
\label{fig:editing} 
\end{figure}

\subsection{Scene Editing and Relighting}

Equipped with the geometry and material decomposition results, we could produce photo-realistic scene images in arbitrary target materials and lighting. As shown in \refFig{fig:editing}, we change the albedo color, metallic, and roughness of the roof, which are then fed to our physically-based render layer and result in significantly different reflectance around the roof area in images. In addition, we synthesize the photo-realistic scene images under different environment lights, which are included in the supplementary.

\Skip{
\begin{figure}[!h] 
\centering 
\includegraphics[width=\linewidth]{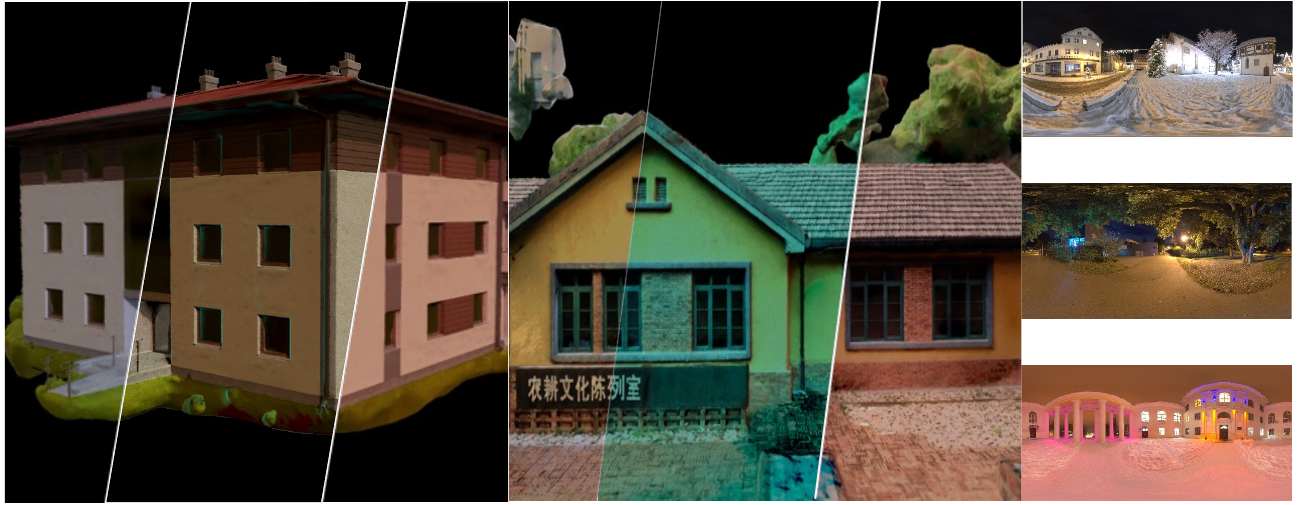}
\caption{Relighting results using real-world captured environment maps. }
\label{fig:relighting} 
\end{figure}

}

%% file: tables/comparsion_syn.tex
\begin{table}
\centering
\setlength{\tabcolsep}{2.0mm}
\caption{Quantitative comparison results of various methods. The inferred images are rendered from reconstructed neural representations.}
\begin{tabular}{c|cccc}
        \Xhline{1.5pt}
                    & \multicolumn{4}{c}{PSNR↑}   \\
        \Xhline{0.8pt}	
                    & Nvdiffrec		& NeRF-OSR		& TensoIR		& \textbf{Ours}		\\
	Building 1	&27.83	& 18.34	& \cellcolor{magenta!60}32.34		& \cellcolor{pink}{29.85}	\\
	Building 2	&30.29	& 19.31			&\cellcolor{pink} 31.21		& \cellcolor{magenta!60} {32.64}	\\
    Building 3	& 32.5		& 21.03			& \cellcolor{magenta!60}34.99		& \cellcolor{pink}{32.74}\\
    Museum	& 17.32		& 14.37			& \cellcolor{pink}{20.88}		& \cellcolor{magenta!60}{21.81}\\
    Villa	& 22.94		& 12.75		& \cellcolor{pink}26.73		& \cellcolor{magenta!60}{27.44}\\
        \Xhline{1.5pt}
\end{tabular}
 \label{tab:delit compare}   
\end{table}

	

%% file: sections/conclusion.tex
\section{Conclusion}
We present a novel inverse rendering framework that enables reconstructing lighting, geometry, and material properties for the facade of the outdoor scene from multi-view aerial images. Our approach represents the scene as neural implicit signed distance with multi-resolution feature grids and models the diffuse color and specular color of appearance with network fields separately. Owing to the effective neural representation, we reconstruct the high-quality geometry of the scene. To further decompose the material properties of any points in the scene surface, we parameterize the spatially-varying material properties of the scene as a neural field and propose an adaptive material segmentation and cluster approach to regularize it. In the meantime, we introduce to represent the environment light with an analytical daylight model. Finally, we jointly optimize the lighting, visibility, and material by a differentiable Monte Carlo render layer to produce photorealistic re-rendering results. In addition, we propose a dataset containing multiview images via aerial 3d scanning and the corresponding Lidar-captured GT point cloud. We demonstrate that our approach is able to achieve SOTA inverse rendering results, outperforming previous neural methods in terms of reconstruction quality, in both synthesized data and real data.